\documentclass[10pt,letterpaper]{article}

\usepackage{arxiv}

\usepackage{pslatex}
\usepackage{apacite}
\usepackage{natbib}
\usepackage{float} 
\usepackage{graphicx}
\usepackage{amsmath}
\usepackage{wrapfig}

\usepackage[utf8]{inputenc} 
\usepackage[T1]{fontenc}    
\usepackage{hyperref}       
\usepackage{url}            
\usepackage{booktabs}       
\usepackage{amsfonts}       
\usepackage{nicefrac}       
\usepackage{microtype}      
\usepackage{lipsum}		
\usepackage{doi}

\newcommand\eg{\textit{e.g.,~}}
\newcommand\ie{\textit{i.e.,~}}

\title{Multimodal Input Aids a Bayesian Model of Phonetic Learning}
 
\author{{\large \bf Sophia Zhi (szhi@mit.edu)} \\
  MIT Department of Computer Science, 50 Vassar St \\ Cambridge, MA 02139
  \AND {\large \bf Roger Levy (rplevy@mit.edu)} \\
  MIT Department of Brain and Cognitive Sciences, 43 Vassar St \\ Cambridge, MA 02139
  \AND {\large \bf Stephan Meylan (smeylan@mit.edu)} \\
  MIT Department of Brain and Cognitive Sciences, 43 Vassar St \\ Cambridge, MA 02139}

\begin{document}
\maketitle

\begin{abstract}
One of the many tasks facing the typically-developing child language learner is learning to discriminate between the distinctive sounds that make up words in their native language.
Here we investigate whether multimodal information---specifically adult speech coupled with video frames of speakers' faces---benefits a computational model of phonetic learning.
We introduce a method for creating high-quality synthetic videos of speakers' faces for an existing audio corpus.
Our learning model, when both trained and tested on audiovisual inputs, achieves up to a 8.1\% relative improvement on a phoneme discrimination battery compared to a model trained and tested on audio-only input.
It also outperforms the audio model by up to 3.9\% when both are tested on audio-only data, suggesting that visual information facilitates the acquisition of acoustic distinctions.
Visual information is especially beneficial in noisy audio environments, where an audiovisual model closes 67\% of the loss in discrimination performance of the audio model in noise relative to a non-noisy environment.
These results demonstrate that visual information benefits an ideal learner and illustrate some of the ways that children might be able to leverage visual cues when learning to discriminate speech sounds.

\textbf{Keywords:} 
first language acquisition; phonemic category learning; multimodal learning; Bayesian cognitive modeling; vision impairment
\end{abstract}

\section{Introduction}

As hearing infants are exposed to speech, they learn what distinguishes the acoustic signatures of the words that make up their language \citep{WERKER1984}.
For example, English-learning infants learn to differentiate /l/ and /r/ such that they can differentiate words like `lock' and `rock,' while Japanese-learning infants do not learn to differentiate these sounds, as they are not used contrastively in Japanese \citep{Kuhl2006}.
Ample experimental research demonstrates that infants are sensitive to acoustic distributional information in the speech they hear and that they use this information to learn acoustic distinctions  (\eg \citealp{Maye2002, Maye2008, Yoshida2010}).
How children learn to appropriately discriminate between phonetic categories of their first language---and their manifestations across a variety of different speakers and speech contexts---is an open area of experimental and computational investigation \citep{Maye2002, Werker2007, Antetomaso2017}.

In addition to the information conveyed by sound, the visible movements of a speaker's lips, jaw, tongue, and throat also contain information that can be exploited by an adult listener.
Ample research suggests that adult speech perception is sensitive to visual input reflecting these facial movements \citep{Sumby1954, MacLeod1987, Picou2011, Knowland2016, Bidelman2019, Hardison2003}.
Perhaps most strikingly, audio input corresponding to one phoneme (/b/) combined with visual input corresponding to a second phoneme (/g/) causes listeners to perceive a different, third phoneme (/d/) (the McGurk effect, \citealp{MCGURK1976}).
More recent work has shown that infants pay visual attention to speakers' mouths while acquiring their first language \citep{HillairetdeBoisferon2016, HillairetdeBoisferon2018, Lewkowicz2012}, and access to speaker faces enhances phoneme discrimination among 6-month-old infants \citep{Teinonen2008}.
Further, visual attention to speaker's mouths for 12-month-old infants has been found to correlate with productive vocabulary outcomes at 18 and 24 months \citep{TENENBAUM2014}.
This effect is corroborated by the 7.5-month average vocabulary delay observed in a sample of visually impaired and blind children compared to typically-developing sighted children \citep{Campbell2024}.\footnote{Visually impaired and blind children eventually obtain the same vocabulary, along with similar linguistic competence \citep{landau1985language}.
Some evidence even points to processing advantages among blind individuals, \eg \citet{loiotile2020enhanced}.}
Taken together, these studies suggest that multimodal input plays a role in the acquisition of phonetic knowledge and subsequent phonetic processing in typically-developing children. 

\begin{figure*}
    \centering
    \includegraphics[width=\textwidth]{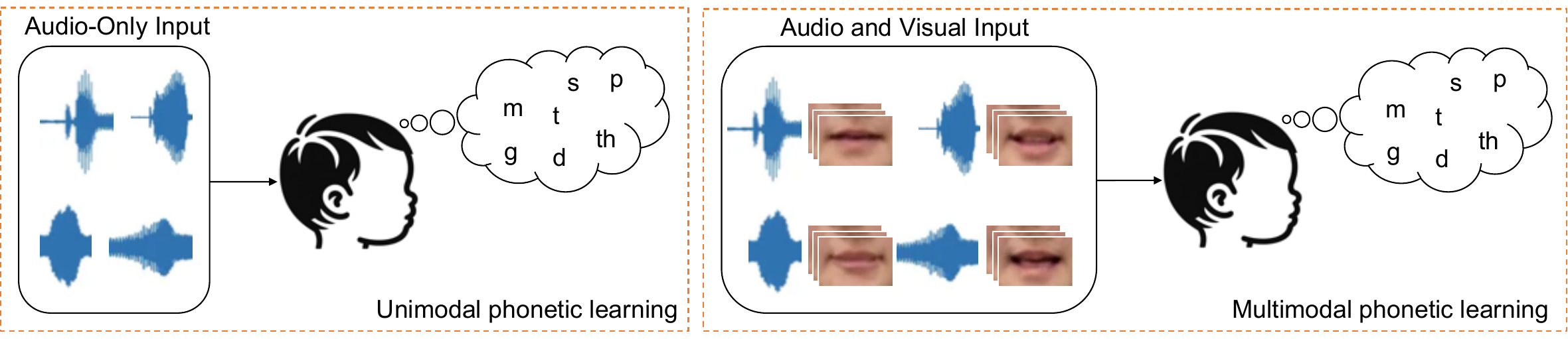}
    \caption{Two hypotheses about what matters for the acquisition of phonetic categories: Under the first or ``unimodal'' hypothesis, children learn from and use the acoustic signal alone; under the second ``multimodal'' hypothesis, child learners also make use of the visual information from the speaker's articulators.}
    \label{fig:intuition}
\end{figure*}

However, multimodal input has not played a prominent role in computational modeling of sound category learning.
Most computational approaches in the acquisition literature have assumed that infants learn these categories through some form of distributional learning of purely acoustic features \citep{deBoer2003, Vallabha2007, Feldman2013, Dillon2013, Frank2014, Mcmurray2009,Feldman2013}.
Consistent with this audio-only approach to modeling, some children \textit{must} learn phonetic contrasts from the audio input alone (i.e., visually impaired children), however this leaves open the possibility that typically developing children learn by leveraging both the acoustic and visual signal.

In contrast to the child language acquisition literature, researchers in automatic speech recognition (ASR) have often attempted to use the visual signals from speakers' faces to improve word recognition.
\citet{Coen2006} focused specifically on speaker faces by modeling vowel acquisition as unsupervised clustering of lip position measurements and vowel formants, but was limited to controlled, vowel-only data, and provided no quantitative evaluation.
\citet{Chaabouni2017} found that a semi-supervised neural network trained on a lip-reading dataset (\citealp{benezeth2011bl}) benefited from adding visual information in discriminating certain phonemic contrasts, but did not demonstrate overall improved performance.
More successfully, \citet{Wang2016} found that a Hidden Markov Model trained on audiovisual data outperformed its audio-only baseline on word recognition, especially in relatively high-noise audio.
\citet{Afouras2020} achieved state-of-the-art word error rates (WER) on two lip-reading benchmarks using a lip-reading model distilled from an ASR model that was trained on audio-only input.
\citet{Chen2023} also demonstrated improved WER with audiovisual data over audio-only and visual-only data using a reinforcement learning-based model, with the relative improvement of audiovisual data improving as noise increases.

A broader body of work in ASR has investigated how visual input of all sorts---not just speaker's faces, but also images of scenes---can help phoneme discrimination and word recognition.  
Using a dataset of audio captions coupled with scene images, \citet{Harwath2020} introduced an audiovisual convolutional network that greatly improved the phoneme discrimination error rate compared to an audio-only baseline.
On a similar dataset, \citet{Srinivasan2020} found that using the visual modality improved an ASR model's WER as well as the model's ability to recover masked words.
Other neural audiovisual representation learning models, also trained on images with audio captions, did not result in better discrimination performance than their audio-only baselines, although they did improve performance on a semantic task \citep{Alishahi2021, Peng2022}.
With instructional action-based videos, incorporating utterance-level \citep{Miao2016} or video-level \citep{Ghorbani2021} visual features has been found to improve upon the WER of corresponding baseline ASR models.
While these algorithms look at a broader swath of visual input, they could be adapted to look specifically at speakers' lips and mouths.

In this paper we present an implemented computational model of unsupervised multimodal phonetic acquisition and processing from naturalistic speech.
Our model shows clear benefits from visual input over the established unimodal baseline model of \citet{Schatz2021}, which we extend to handle multimodal input by coupling audio recordings with video frames of a speaker's mouth.
Following \citet{Schatz2021}, we use a nonparametric Bayesian approach to cluster short (25 ms) acoustic windows, and evaluate the trained model's ability to appropriately discriminate between recordings with different phonemic identities by evaluating the similarity of recordings in an embedding space derived from sequences of cluster identities. 
Consistent with the logic of \citet{Schatz2021}, we do \textit{not} presuppose that children's knowledge of sounds is organized around phonemes, but instead assume that evaluating a model's ability to group examples of phonemes as a key way to evaluate the efficacy of a learning system, even if that learning system is organized around other units of representation.
For example, the \citet{Schatz2021} model sorts short acoustic intervals (25ms) into a much larger number of clusters than is typical of phonemic inventories, yet is nonetheless able to perform well on a phonemic discrimination task.
As a test of the model's abilities to \textit{process} phonetic distinctions, we evaluate whether the model trained with visual information performs better when visual information is available at test time.
As a test of the persistent contribution of visual information to phonetic category \textit{learning}, we additionally evaluate whether the advantage of audiovisual training remains when auditory-only input is presented to the model at test time. 
Finally, to test whether visual input confers robustness to phonetic processing in noisy environments, we also test the model's performance when it only has access to noisy auditory input at test time.

\begin{figure*}[t!]
    \centering
    \includegraphics[width=\linewidth,trim={0 22mm 0 0},clip]{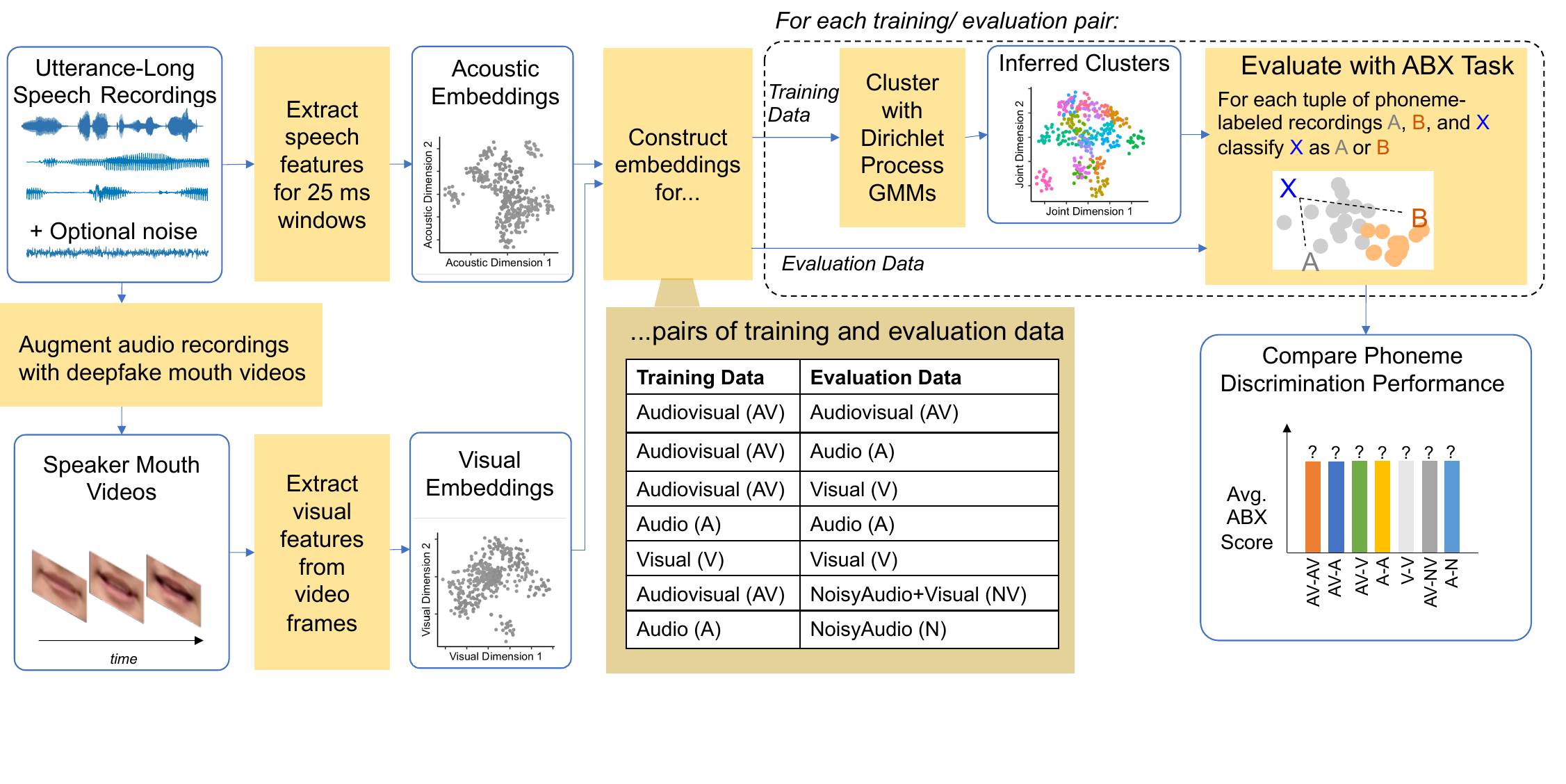}
    \caption{Overview of dataset creation, derivation of acoustic and visual embeddings, clustering, and evaluation. We vary the modalities used at both train and test time and then test the resulting models' phoneme discrimination performance.}
    \label{fig:overview}
\end{figure*}

\section{Approach}

Our modeling approach follows \citep{Schatz2021} in how it handles acoustic input:
extracting acoustic features, clustering these features to produce distributions over a lower-dimensional acoustic space, and then comparing the similarity of new tokens in this lower dimensional space in a phoneme discrimination task. 
In addition to the acoustic input, however, our approach incorporates visual input as well (Fig.~\ref{fig:overview}).
The Bluelips dataset \citep{Bachman2011} used by \citet{Chaabouni2017} contains only a highly restricted set of example sentences, includes many different speakers, and suffers from low video quality (including many compression artifacts). 
In order to fill a crucial dataset gap for multimodal visual–facial--auditory phonetic acquisition, we create a high-quality synthetic video dataset (of ``deepfakes'' of adult faces).
From a pretraining subset of these videos, we learn the parameters used for video feature extraction without making use of any phonetic category information, and concatenate these visual embeddings with the audio embeddings.
We then perform unsupervised clustering of these multimodal embeddings using a Dirichlet Process Gaussian Mixture Model (DPGMM; \citealp{Chen2015}), which  learns the number of clusters as well as the mean and variance of each cluster.
We evaluate the model in its ability to discriminate between phonemes in an ABX task, which measures how often the model finds a speech recording X to be more similar to a recording A of the same phoneme rather than recording B of a different phoneme.
The ABX battery comprises every possible ABX triple in a test set composed of a random subset of videos from the same source as the training set.


To measure the impact of different information sources, we evaluate several combinations of modalities at training time and test time (Figure \ref{fig:overview}) and test seven key comparisons (Table \ref{train-test_comparisons}).
In each case, we expect the first train/test combination to outperform the second.
\textbf{Comparison 1}, AV-AV vs A-A, measures the benefit of using visual information in phoneme discrimination at test time.
\textbf{Comparison 2}, AV-A vs A-A, measures the benefit of visual information in phonemic category learning, when no visual information is available at test time.
\textbf{Comparison 3}, AV-V vs V-V,  measures the benefit of auditory information with respect to a visual-only baseline (\ie the utility of auditory information for learning visual representations for lip-reading).
\textbf{Comparison 4}, AV-NV vs A-N, measures how much visual information may aid speech processing when the audio signal is presented in noise.
We present three more comparisons as sanity checks: \textbf{Comparison 5}, A-A vs V-V, \textbf{Comparison 6}, AV-AV vs AV-NV, and \textbf{Comparison 7}, A-A vs A-N.
We leave the evaluation models trained on noisy audio to future work.

\begin{table}
    \centering
    \begin{tabular}{c|c}
        (1) & AV-AV $>$ A-A \\
        (2) & AV-A $>$ A-A \\
        (3) & AV-V $>$ V-V \\
        (4) & AV-NV $>$ A-N \\
        (5) & A-A $>$ V-V \\
        (6) & AV-AV $>$ AV-NV \\
        (7) & A-A $>$ A-N \\
    \end{tabular}
    \caption{Predicted performance in key model comparisons }
    \vspace{-3mm}
    \label{train-test_comparisons}
\end{table}


\subsection{Dataset}

To best mimic the conditions under which infants learn phonemic categories, a dataset with child-directed (or at least child-available) speech such as CHILDES \citep{MacWhinney1992} would be optimal.
However, few if any CHILDES transcripts include high quality videos of caregiver faces, and they reflect a wide variety of lighting conditions and angles.
As an additional concern, the ABX evaluation task requires the use of phoneme-level aligned transcriptions to extract recordings for the evaluation, which CHILDES does not have. Therefore we use the Buckeye corpus \citep{Pitt2007a}, which includes approximately 40 hours of audio recorded in 2000 from 40 English speakers from Ohio speaking spontaneously to an interviewer. 
Each audio file is labeled with the sex of the speaker, age group of the speaker, and sex of the interviewer.
This corpus also includes word- and phoneme-level transcriptions, which were first generated from an automatic aligner, but later hand-corrected by trained researchers. 

\subsubsection{Deepfake generation}
We augment the Buckeye corpus by generating a set of synthetic videos of moving faces, using the deepfake generation model of \citeauthor{Prajwal2020} (\citeyear{Prajwal2020}; see left side of Fig.~\ref{fig:overview}).
First, we filmed a short video of a speaker (the first author) facing the camera and producing a dummy sentence with normal affect.
We then used this base video to generate all deepfakes, so that the resulting deepfakes had minimal variation in speaker identity, head position, and lighting.
To more closely simulate the typical infant's experience of learning primarily from a relatively small number of speakers, we used all and only Buckeye audio files labeled as female speaker, young speaker, and female interviewer, resulting in five independent speakers. 
The labels are limited to these categories to reduce variance between different speakers and in the sounds labeled as non-phonemic and produced by the interviewer.
These audio files are then split into shorter utterances (appropriate for generating deepfakes) by splitting audio files on any segment with a non-phonemic label such as silence or vocal noise.\footnote{
The choice of how to split audio recordings into utterances is potentially consequential in that segmentation of recordings determines the audio input to the deepfake model. In future work we will investigate more sophisticated segmentation strategies that more accurately reflect utterance boundaries.
}
The final deepfakes have 60 video frames per second of 960x540 pixels and 8-bit color depth, and audio recorded at 16 kHz.
In total, this dataset had 0.8 hours of video in the pretraining set, 0.67 hours of video in the training set, and 0.33 hours of video in the evaluation set. 


\subsection{Model}
\subsubsection{Audio features}
The audio features are Mel Frequency Cepstrum Coefficients, or MFCCs \citep{Jurafsky2014}, and their first and second time derivatives, following the procedure in \citet{Schatz2021}. 
The input audio is segmented into overlapping 25 ms windows of time spaced 10 ms apart.
The time-domain audio signal from each window is then used to calculate 13 MFCCs, from which the first and second time derivatives can be estimated using the preceding and following 3 windows.
For the noisy-audio evaluation cases, we added Gaussian noise at a signal-to-noise ratio of 5 to the audio recordings before computing MFCCs.

\subsubsection{Video features}
\begin{figure}[t]
    \centering
    \includegraphics[width=.5\linewidth]{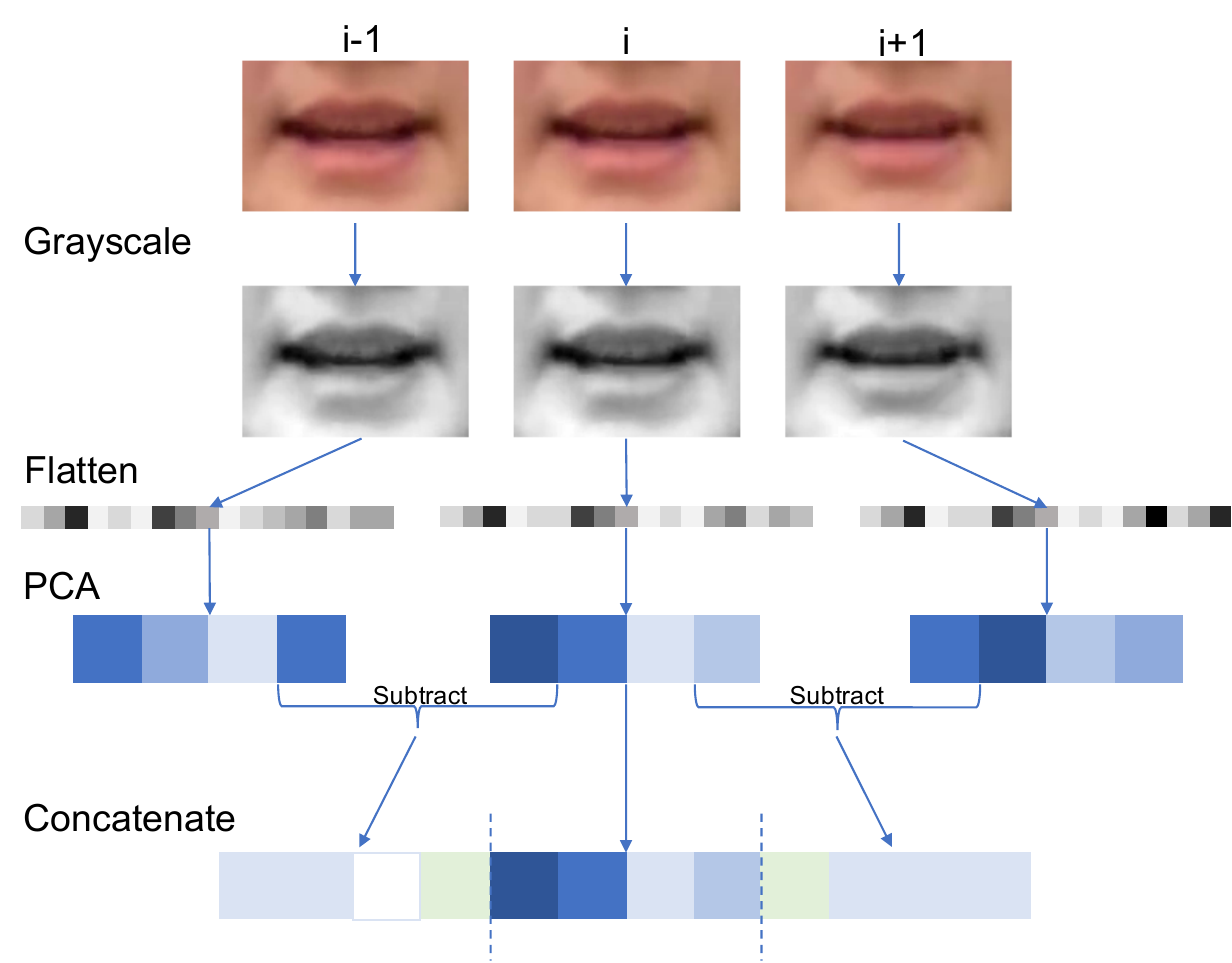}
      \vspace{-5mm}
    \caption{Video feature extraction. Each frame is transformed to a low-dimensional representation, and a set of video features that captures both static and dynamic information is derived by concatenating 1), the middle frame's representation with 2), the difference between that and the previous frame's representation and with 3), the difference between the next frame's representation and the middle frame's representation.}
    \label{fig:video_features}
    \vspace{-5mm}    
\end{figure}

Video features are taken over the same 25ms windows as the audio features, for consistency.
For each window, the closest video frame that occurs before the center timestamp of the window, as well as its two adjacent frames, are transformed into a relatively low-dimensional representation of the mouth by cropping to a 100x150 box around the mouth region and converting from color to 8-bit grayscale (Fig. \ref{fig:video_features}).
Because the speaker head is relatively still in the deepfakes, the mouth bounding box was defined by hand in advance and is the same for each video frame.
The grayscaled image is calculated as $0.2989r+0.587g+0.114b$, where $r,g,b$ are the red, green, and blue channels of the input image, respectively \citep{Yang2021}. 
Then we reduce the dimensionality of the visual input from each frame by computing the coefficients of a linear combination of ``eigenmouths," which are a set of pictures (represented as vectors) that can be factored using Principal Components Analysis (PCA) into a set of vectors that can be recombined with weightings to approximate the space of possible mouth images (following the ``eigenfaces" approach for facial recognition; \citealp{Sirovich1987, Turk1991}).
To capture both the static information of what the mouth looks like as well as dynamic information of how the mouth is moving, the video features for a single window are the concatenation of the eigenmouth coefficients of the center video frame with the differences between the center frame's coefficients and the adjacent frames' coefficients.

\vspace{-3.5mm}
\subsubsection{Pretraining}
A pretraining set, consisting of held-out deepfakes from the same source as the training and evaluation sets, is used to determine the PCA parameters that represent the eigenmouth coefficients.
We match each acoustic window with the closest video frame, crop the image to the speaker's mouth, and convert the image to grayscale.
To select the number of PCA components to retain for video frames, we trained a video-only DPGMM on a small subset of videos and then evaluated on a small ABX battery while varying the number of dimensions.
We selected the number of PCA components that produced the best video-only ABX performance: 4 components achieved an average ABX score of 0.705.

\subsubsection{Clustering model}
Following \citep{Schatz2021}, we use a Dirichlet Process Gaussian Mixture Model \citep{Chen2015} to cluster the audiovisual embeddings without supervision into multivariate Gaussian clusters. 
The number of clusters in a DPGMM is learned from the training data; the number of clusters is partially governed by a hyperparameter $\alpha$ which represents the weight given to a new cluster for each datapoint seen in sequence.
With the first $n-1$ points forming $K$ clusters, the distribution over clusters for the $n$th point is:
\[
  X_n | X_{n-1}...X_1 :
  \begin{cases}
    \text{cluster } K=1 & \text{with } p = \frac{\alpha}{n-1+\alpha} \\
    \\[1pt]
    \text{cluster } k & 
    \!\begin{aligned}
        & \text{with } p = \frac{n_k}{n-1+\alpha} \\
        & \text{for } k=1,...,K
    \end{aligned}
  \end{cases}
\]
Consequently, $\alpha$ can be increased to push the model to find more clusters, though the exact number of clusters is determined by fit to the data.
We use the DPGMM inference procedure implemented by \citet{Chang2013} and the same hyperparameters as \citet{Schatz2021}, including setting $\alpha=1$, 1500 iterations of training, and initializing the model with 10 clusters.
During training, the DPGMM optimizes the number of clusters, mean and variance of each cluster, and assignment of each instance in the training set to clusters.
Directly optimizing all parameters is intractable, so a highly parallelizable version of Gibbs sampling is used to learn the parameters efficiently \citep{Chang2013}.
Once trained, the DPGMM takes a feature vector associated with a single window as input and outputs a probability distribution over cluster identities.

\subsubsection{Input representations for models}
The representation of input features differs slightly between each case.
In the AV-AV case, the audio and visual features of each time window are simply concatenated together, both for the purposes of training and testing.
In the single-modality models, only the audio features or only the visual features of each time window are included in the training or test features.
For the audiovisual-audio and audiovisual-video cases, the training features are the concatenation of audio and video features, but the test features are only audio features or only video features, with the missing modality truncated from the representations at test time.
In these cases, given the smaller dimensionality of the test features with respect to the training features, posteriors are calculated by disregarding the cluster parameters that correspond to the unknown dimensions of the test input.

\subsection{Evaluation}
\begin{figure*}
    \centering
    \includegraphics[width=\linewidth,trim={0 3cm 0 0}]{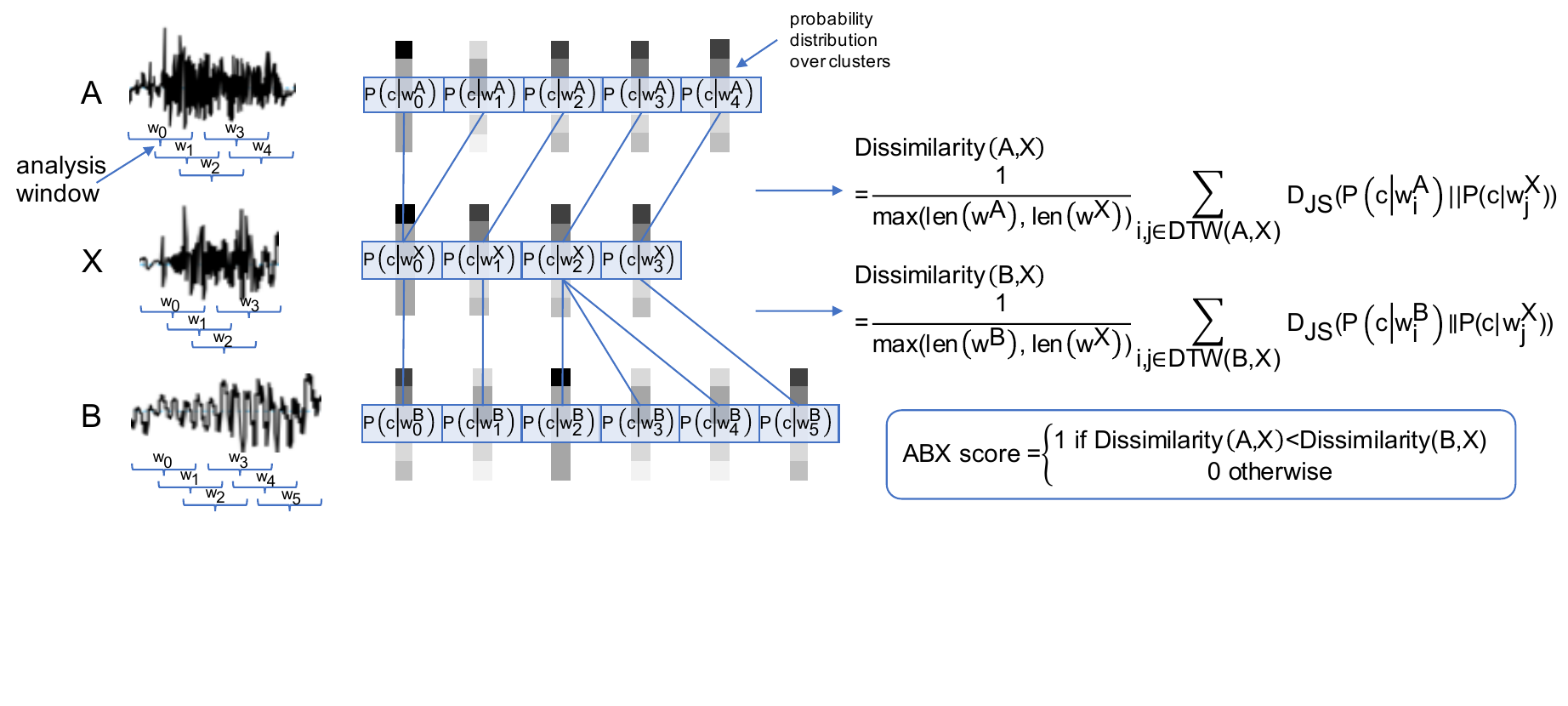}
        \vspace{-10mm}
    \caption{ABX similarity calculation for evaluation. We compare audio/video recordings of potentially different durations using dynamic time warping to match each window in one recording to a window in the other, then calculate the overall dissimilarity by averaging the divergence between each pair of matched windows. $w$ indicates window and $c$ indicates cluster assignment.}
        \vspace{-5mm}
    \label{fig:abx}
\end{figure*}

Each trained DPGMM is evaluated using an ABX discrimination task \citep{Schatz2013}.
An ABX battery collects unique triples of sound tokens $(A,B,X)$ for which $A$ and $X$ have the same phoneme label, $B$ and $X$ have different phoneme labels, and all three sounds occur in the same phonemic context (\ie, before and after the same phonemes, including contexts where an adjacent phoneme belongs to a different word).
Keeping the phonemic context constant is intended to account for coarticulation effects, where the production of a sound is affected by its neighboring sounds \citep{Daniloff1973}.
Each sound token spans the entire time range assigned to its phoneme label, as specified by the phoneme alignments in the Buckeye corpus.
For a single ABX triple, a score of 1 is assigned if the model finds X to be more similar to A than B, 0 if X is more similar to B than A, and 0.5 if X is equally similar to both.

The above-described process of calculating the model's similarity between two sound tokens of potentially unequal lengths is depicted in Figure \ref{fig:abx}.
Each sound token is represented as a sequence of cluster probability distributions ($P(c)$) corresponding to the sequence of windows whose centerpoints occur within the token's start and stop time. 
Following \citet{Schatz2021}, we compare $A$ and $X$ using an optimal matching between the windows of $A$ and the windows of $X$ found via dynamic time warping \citep{Vintsyuk1968}.
Dynamic time warping is a method which aligns two temporal sequences, in this case the windows of $A$ and $X$, by considering all matchings in which each window of $A$ is matched to a window of $X$ and vice versa, and the indices of the matched windows are monotonically increasing along each sequence. The output of dynamic time warping is a set of matched pairs of windows (not necessarily one-to-one).
The optimal matching is that which minimizes the average Jensen-Shannon divergence (\citealp{Kullback1951, Dagan1997}):
\begin{align}
    D_{JS}(P||Q)=\frac{1}{2}\sum_{c\in \mathcal{C}} P(c)\ln{\frac{P(c)}{Q(c)}}+\frac{1}{2}\sum_{c\in \mathcal{C}} Q(c)\ln{\frac{Q(c)}{P(c)}}
    \label{eq:js-divergence}
\end{align}
 between the two probability distributions over the set of clusters $\mathcal{C}$ in each matched pair $(w_i^{A}, w_j^{X})$ resulting from the dynamic time warping alignment.
This minimum average JS divergence serves as a \textit{dis}similarity measure between $A$ and $X$.

The discrimination score for a specific AB contrast is the average score across all triples with phoneme A and phoneme B.
The overall discrimination score is then calculated by averaging over all vowel---vowel contrasts and consonant---consonant contrasts, following \citet{Schatz2021}.
We compute average discrimination results across 10 different fitted DPGMMs for each train/test combination. 

In addition to reporting absolute improvement in each key comparison, we compute relative improvement by taking into account that the ABX task has a chance baseline of 0.5.
For example, the audiovisual-audiovisual case's relative improvement over the audio-audio case is computed by comparing each case's increase in performance with respect to chance:
\begin{equation}
    \text{relative improvement} = \frac{\text{AV-AV score} - \text{A-A score}}{\text{A-A score} - 0.5}
\end{equation}

\section{Results}


\begin{figure}[t]
    \centering
    \includegraphics[width=0.6\linewidth]{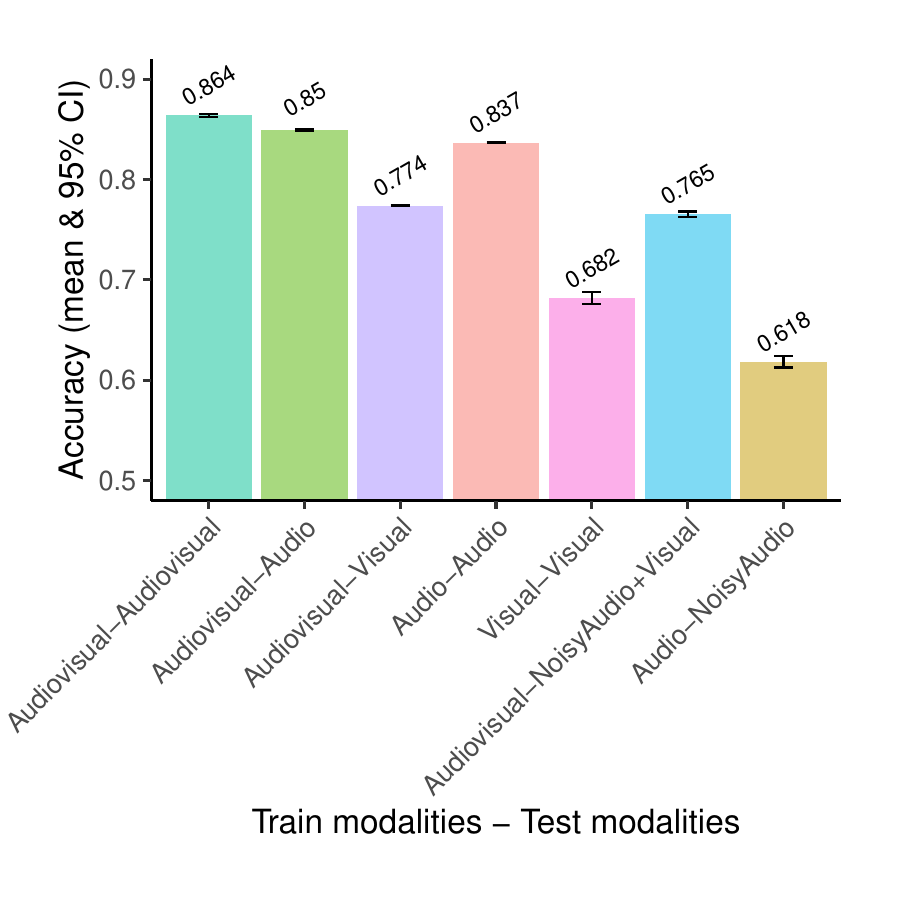}
        \vspace{-7mm}
    \caption{Overall ABX phoneme discrimination score for all combinations of train/test modalities. Results are averaged over 10 fitted DPGMMs and error bars mark 95\% confidence intervals.}
        \vspace{-3mm}
    \label{fig:results_abx}
\end{figure}

\textbf{Comparison 1}: 
We predicted that a model trained and tested on audiovisual data would perform better than a model trained and tested on audio data alone.
Alternatively, visual input might be redundant with the auditory input, or provide distracting information that makes learning phonemic categories more difficult.
Our results support the first hypothesis.
Across datasets, the audiovisual-audiovisual case shows a 2.7\% absolute improvement over the audio-audio case ($U=100, p<10^{-3}$ with a two-tailed Mann-Whitney U-test; Fig. \ref{fig:results_abx}). 
This translates to an 8.1\% relative improvement relative to chance.
This provides evidence in favor of the hypothesis that visual information can indeed be beneficial for phonemic category processing.


\textbf{Comparison 2} 
Our second prediction was that a model trained on audiovisual data would perform better than a model trained on audio data when both models are tested on audio-only data.
This test is stronger than the preceding comparison in that it would mean multimodal information has a lasting impact on the quality of inferred phonemic representations  which transfers to later processing of speech even when visual input is not present.
Our results are consistent with this hypothesis: the audiovisual-audio case shows a 1.3\% absolute and 3.9\% relative improvement over the audio-audio case ($U=100, p<10^{-3}$).

\textbf{Comparison 3} The audiovisual-video case shows an absolute improvement of 9.2\% and a relative improvement of 50.7\% over the video-video case, ($U=100, p<10^{-3}$). 
Symmetrical to Comparison 2, access to acoustic signal at training time allows the model to infer better visual distinctions between phonemes, allowing it to perform better on video-only test data.

\textbf{Comparison 4} 
We predicted that an audiovisual model tested on noisy audio and standard video (AV-NV) would outperform an audio-only model tested on noisy audio (A-N).
As expected, the AV-NV case outperforms the A-N case by a 14.7\% absolute improvement ($U=100,p<10^{-3}$), which represents 67\% of the 21.9\% absolute difference between A-N and A-A.
These results support the hypothesis that visual information can help counteract the difficulty of processing speech in a noisy environment.

\textbf{Comparisons 5-7} As expected, the audio-audio case is also consistently better performing than the video-video case by an absolute 15.5\% and a relative 85.2\% ($U=100, p<10^{-3}$).
Video-video scores are well above chance, suggesting that visual information alone is sufficient for phoneme discrimination in many cases.
Similarly, the audiovisual model performs better when tested on standard audio (AV-AV) than when tested on noisy audio (AV-NV) by an absolute 9.9\% and a relative 27.3\% ($U=100, p<10^{-3}$).
The audio model performs better when tested on standard audio (A-A) than when tested on noisy audio (A-N) by an absolute margin of 21.8\%, corresponding to a relative improvement of 184\% ($U=100, p<10^{-3}$).

\section{Discussion}
\vspace{-1mm}
The results of this work suggest that utilizing visual information from the speaker's mouth can help a learner infer more useful phonetic categories.
These phonetic categories offer the learner better performance in a phoneme discrimination task, especially when the audio is noisy. 
Not only does visual information help at processing time, but the representations learned from the combined audiovisual signal yield more useful acoustic representations that can benefit later speech processing when the visual signal is not available. 
Visual information is especially beneficial in noisy audio environments, where the improvement in phonetic discrimination afforded by the inclusion of the visual signal is much larger than the corresponding difference in the clean audio environment.
These benefits can be explained by the additional information provided by the visual signal. 
Each video frame reflects relatively long-interval gestural information corresponding to many audio frames.
The synthetic videos also contain information not present in the audio signal, as the video frames are not a deterministic function of immediately nearby acoustic windows.
Instead, the video frames reflect a broader range of model-based expectations (i.e., how mouths move) and are informed by relatively long intervals of preceding and following audio.

While this study establishes that visual cues from the speaker's mouth can provide useful, non-redundant information for phonemic category learning, it has several limitations.

One limitation is that this model sees a more constrained set of visual inputs than what typically-developing children would experience.
This model sees only one speaker's mouth, with consistent angle and lighting; further, visual information is always available to the learner.
If a wider range of mouth positions and speaker faces were included, the increased variance not relevant to phonemic distinctions would likely make it more difficult to cluster the video features.
Instead, this approach makes the simplifying assumption that children at the age of phonemic category learning already know how to extract the relevant information from speakers' mouths.

A second limitation of the visual signal is that the synthetic videos in our dataset are not as fine-grained as real human movement.
Some mouth movements in the deepfakes may be simplified or imprecise, making the deepfakes less informative than the visual signal that children receive.
We may have also lost information provided by the visual signal due to the way we reduced the dimensionality of mouth images.
We selected the first $n$ PCA components that resulted in the best discrimination performance, where the components are ordered by the amount of variance explained.
However, the usefulness of each component for phoneme discrimination is not necessarily related to the amount of variance the component explains.
For example, we found that the performance of the first 7 components improves upon the performance of the first 6 components, and thus the 7th PCA component might represent useful information when combined with the first four.
On the other hand, the 5th and 6th components do not appear useful because including them in the video features decreases performance.
A more thorough approach than the one presented here would try to discover which of the combinations of PCA components are most helpful, regardless of the order of variance explained.


Similarly, the audio in this model is an imperfect representation of what children hear. 
The Buckeye corpus is adult-directed, so it lacks many of the qualities of infant-directed speech such as hyperarticulated vowels \citep{Kuhl1997, Cristia2014} and exaggerated prosody \citep{Fernald1984, Fernald1989}.
It is possible (though debated) that infant-directed speech facilitates children's learning of phonemic distinctions \citep{Cristia2014, Eaves2016}, thus we might expect higher overall performance if we were able to train on a corpus of child-directed speech.

Our model is also trained on a very small sample of speech---much smaller than what hearing infants typically hear---so children may be more effective at using both modalities when learning to process phonetic input.
We also note that the current study uses a smaller sample of training data than \citet{Schatz2021}, which likely contributes to the difference between the ABX scores found in their work and those found in this work (another factor may be different preprocessing methods).

Given our smaller dataset, it would also be worthwhile to test in future work newer acoustic models known to perform well at phonemic discrimination, such as wav2vec \citep{Schneider2019} or contrastive predictive coding \citep{Oord2019}. With stronger acoustic models, we may find a diminished benefit of adding visual information.

\section{Conclusion}
\vspace{-1mm}
We present a Bayesian model of phonemic category learning and describe a method to generate high-quality videos of mouth regions using recent deepfake techniques.
We find that an audiovisual model tested on audiovisual input has better phoneme discrimination performance than an audio model tested on audio input, suggests that visual information is useful for phonetic processing when used during learning and processing.
Our second finding, that an audiovisual model performs better than an audio model when both are tested on audio input, suggests a lasting benefit of visual information, which 
yields better acoustic representations that facilitate phoneme identification even when visual information is not available.
Third, we see that visual information confers robustness on the interpretation of noisy audio. 
The results in this work speak to the possibility that visual information from speakers' mouths is useful for phonetic processing both when visual input is available as well as when it is not.

\clearpage

\bibliographystyle{apacite}
\footnotesize
\setlength{\bibleftmargin}{.125in}
\setlength{\bibindent}{-\bibleftmargin}

\bibliography{CogSci_Template}

\end{document}